\documentclass[sigconf]{acmart}
\AtBeginDocument{%
  }

\setcopyright{acmlicensed}
\copyrightyear{2018}
\acmYear{2018}
\acmDOI{XXXXXXX.XXXXXXX}
\acmConference[Conference acronym 'XX]{Make sure to enter the correct
  conference title from your rights confirmation email}{June 03--05,
  2018}{Woodstock, NY}
\acmISBN{978-1-4503-XXXX-X/2018/06}

\begin{document}

\title{Enhancing AI-Based Tropical Cyclone Track and Intensity Forecasting via Systematic Bias Correction}


\author{Peisong Niu}
\authornotemark[1]
\email{niupeisong.nps@alibaba-inc.com}
\affiliation{
  \institution{DAMO Academy, Alibaba Group}
  \city{Hangzhou}
  \country{China}
}

\author{Haifan Zhang}
\authornote{Authors contributed equally to this research.}
\email{zhanghaifan.zhf@alibaba-inc.com}
\affiliation{
  \institution{DAMO Academy, Alibaba Group}
  \city{Hangzhou}
  \country{China}
}

\author{Yang Zhao}
\authornotemark[1]
\email{zhaoy2024@smail.nju.edu.cn}
\affiliation{
  \institution{State Key Laboratory of Severe Weather Meteorological Science and Technology, Nanjing University}
  \city{Nanjing}
  \country{China}
}

\author{Tian Zhou}
\authornotemark[1]
\email{tian.zt@alibaba-inc.com}
\affiliation{%
  \institution{DAMO Academy, Alibaba Group}
  \city{Hangzhou}
  \country{China}
  \postcode{}
}

\author{Ziqing Ma}
\email{maziqing.mzq@alibaba-inc.com}
\affiliation{%
  \institution{DAMO Academy, Alibaba Group}
  \city{Hangzhou}
  \country{China}
}

\author{Wenqiang Shen}
\email{wqshen91@163.com}
\affiliation{%
  \institution{Zhejiang Meteorological Observatory, China Meteorological Administration}
  \city{Hangzhou}
  \country{China}
}

\author{Junping Zhao}
\email{jpzhaozj@qq.com}
\affiliation{%
  \institution{Zhejiang Meteorological Observatory, China Meteorological Administration}
  \city{Hangzhou}
  \country{China}
}

\author{Huiling Yuan}
\email{yuanhl@nju.edu.cn}
\affiliation{
  \institution{State Key Laboratory of Severe Weather Meteorological Science and Technology, Nanjing University}
  \city{Nanjing}
  \country{China}
}

\author{Liang Sun}
\email{liang.sun@alibaba-inc.com}
\authornote{Corresponding authors.}
\affiliation{
  \institution{DAMO Academy, Alibaba Group}
  \city{Bellevue}
  \country{USA}
}









\renewcommand{\shortauthors}{Trovato et al.}


\begin{abstract}
    Tropical cyclones (TCs) pose severe threats to life, infrastructure, and economies in tropical and subtropical regions, underscoring the critical need for accurate and timely forecasts of both track and intensity. Recent advances in AI-based weather forecasting have shown promise in improving TC track forecasts. However, these systems are typically trained on coarse-resolution reanalysis data (e.g., ERA5 at $0.25^\circ$), which constrains predicted TC positions to a fixed grid and introduces significant discretization errors. Moreover, intensity forecasting remains limited especially for strong TCs by the smoothing effect of coarse meteorological fields and the use of regression losses that bias predictions toward conditional means.
    To address these limitations, we propose BaguanCyclone, a novel, unified framework that integrates two key innovations: (1) a probabilistic center refinement module that models the continuous spatial distribution of TC centers, enabling finer track precision; and (2) a region-aware intensity forecasting module that leverages high-resolution internal representations within dynamically defined sub-grid zones around the TC core to better capture localized extremes. 
    Evaluated on the global IBTrACS dataset across six major TC basins, our system consistently outperforms both operational numerical weather prediction (NWP) models and most AI-based baselines, delivering a substantial enhancement in forecast accuracy. Remarkably, BaguanCyclone excels in navigating meteorological complexities, consistently delivering accurate forecasts for re-intensification, sweeping arcs, twin cyclones, and meandering events. The system has been fully operational at the Zhejiang Meteorological Observatory, China Meteorological Administration (CMA) during the 2025 Western Pacific typhoon season. It provided real-time forecasts that critically underpinned emergency response, public warnings, and maritime safety operations. This work represents a significant step toward the operational integration of physically informed high-fidelity deep learning models in high-stakes meteorological applications. Our code is available at https://github.com/DAMO-DI-ML/Baguan-cyclone.

\end{abstract}



\received{20 February 2007}
\received[revised]{12 March 2009}
\received[accepted]{5 June 2009}


\newcommand{\model}{\textsc{BaguanCyclone}}
\maketitle

\section{Introduction}
Tropical cyclones (TCs)~\cite{emanuel2003tropical} represent one of the most hazardous weather phenomena on Earth, routinely causing catastrophic damage to life, infrastructure, and economies across tropical and adjacent subtropical regions. Accurate and timely prediction of TC tracks and intensity forecasts remains a cornerstone of operational meteorology, directly influencing evacuation planning, maritime safety, emergency response coordination, and insurance risk assessment. Numerical weather prediction (NWP)~\cite{buizza2018development} models have been long served as the backbone of operational forecasts and research, predicting weather through explicit physical equations. However, the practical deployment of NWP suffers from incomplete comprehension for certain atmospheric physical processes, alongside high computational costs and model spin-up~\cite{warner2023assessing}.

In recent years, AI-based weather forecasting models~\cite{niu2025utilizing, pangu_nature, Fuxi_nature, Chen2023FengWuPT, graphcast, Bodnar2024AuroraAF} have emerged as promising alternatives or complements to NWP. A growing body of literature demonstrates that deep learning architectures can outperform traditional NWP systems in predicting TC trajectories. Although these AI models demonstrate substantial reductions in track forecast errors at extended lead times, highlighting their potential for operational use, their performance within the 72-hour window still lags behind that of NWP~\cite{zhong2024fuxi}.
Furthermore, a fundamental, yet often overlooked constraint remains inherent to nearly all TC forecasting systems that rely on AI-based global weather models:
they are trained exclusively on coarse-resolution reanalysis data, most commonly ERA5 at $0.25^\circ$. As a result, \textbf{the predicted TC positions are inherently constrained to this fixed grid, limiting the spatial fidelity of forecasts and introducing discretization errors} that can be significant relative to the scale of TC motion. 
Despite the emergence of specialized models~\cite{zhong2024fuxi, huang2025benchmark}, specifically designed for extreme events, they still underperform compared to HRES within the critical 72-hour forecast window.
To address this resolution bottleneck, we propose a novel \textbf{probabilistic correction framework} that explicitly models the continuous spatial distribution of likely TC center locations. By computing the expectation of this distribution, we recover TC position estimates at arbitrary spatial precision effectively.

Beyond track prediction, accurate forecasting of TC intensity, typically quantified by maximum sustained wind speed, remains a persistent challenge for both numerical and data-driven models. While recent AI-based systems have shown promise in capturing large-scale steering flows, they consistently underperform in predicting rapid intensification or decay events~\cite{huang2025benchmark}. This limitation stems from two interrelated factors. First, \textbf{the widely used ERA5 reanalysis at $0.25^\circ$ resolution provides only grid-averaged meteorological fields}, which inherently smooth out the sharp gradients and localized extremes characteristic of TC cores.
Second, \textbf{most deep learning models for TC forecasting are trained with regression losses such as mean squared error (MSE) or mean absolute error (MAE), which inherently bias predictions toward the conditional mean of the target distribution}. 
To overcome these limitations, we introduce a \textbf{region-aware intensity correction algorithm} that operates on dynamically defined sub-grid zones surrounding the TC center. Within each region, high-resolution proxies derived from the model’s internal representations are aggregated and calibrated against historical intensity statistics.

\begin{figure}[h]
    \centering
    \includegraphics[width=1\linewidth]{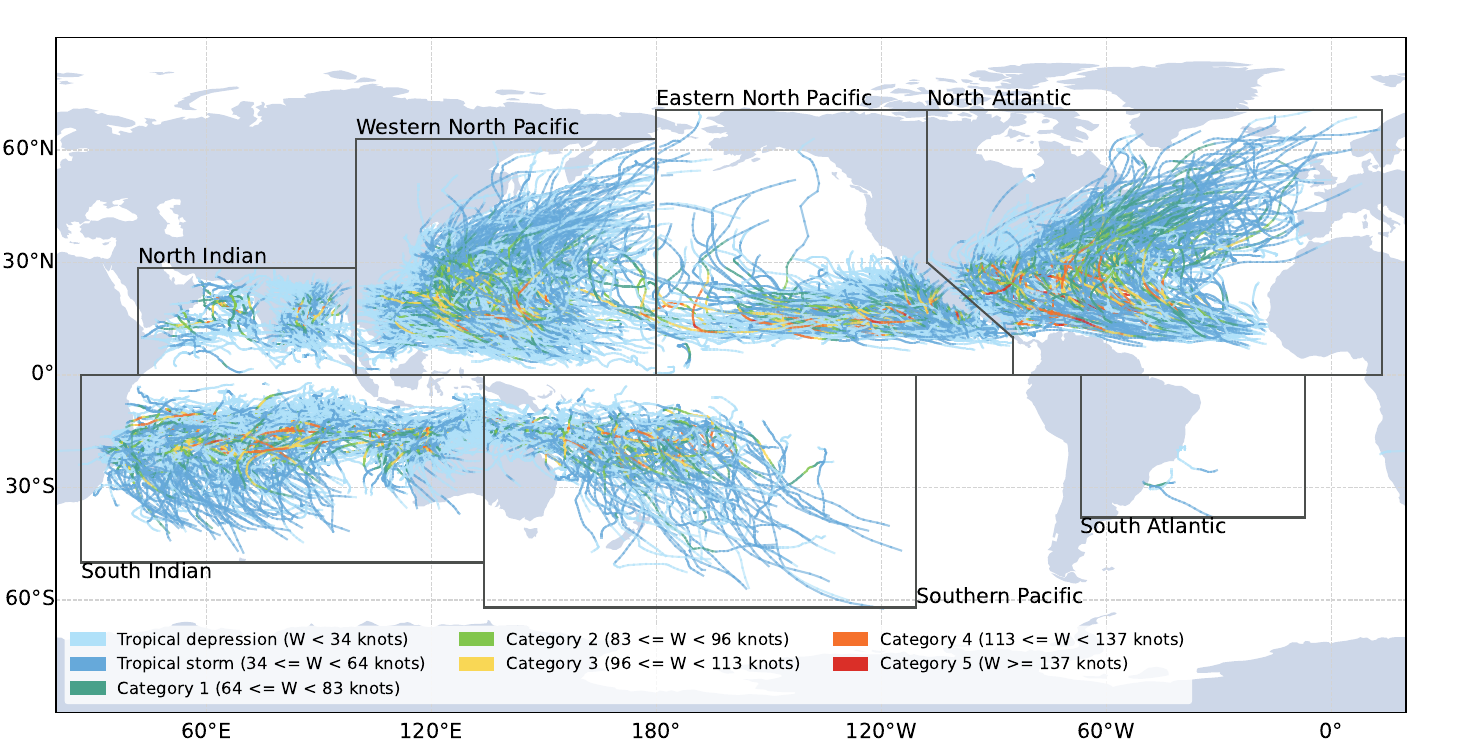}
    \caption{
    Tropical cyclone intensities and tracks. IBTrACS data (2000–2024) colored by Saffir-Simpson category.
    }
    \label{fig:intro}
\end{figure}

To enable holistic TC forecasting, we further \textbf{couple the probabilistic center refinement module with the region-aware intensity correction module into a unified system}. 
This integrated architecture ensures coherence between predicted locations and local structures while streamlining the handling of complex real-world TCs. Consequently, BaguanCyclone effectively models both standard sweeping-arc tracks and intricate scenarios, such as re-intensification, multi-TC interactions, and meandering, that typically defeat conventional numerical methods.
We rigorously evaluate our system on the IBTrACS (International Best Track Archive for Climate Stewardship) dataset, the global standard for historical TC records. 
Experiments across six major TC-prone basins, including the Western North Pacific (WP), Eastern North Pacific (EP), North Atlantic (NA), North Indian Ocean (NI), South Indian Ocean (SI) and South Pacific (SP) (Fig.~\ref{fig:intro}), demonstrate consistent and significant improvements over both operational NWP models and AI-based baselines. Note that we skip the South Atlantic (SA) basin from our analysis due to its insufficient number of tropical cyclone events.

More importantly, the system has been operationally deployed at Zhejiang Meteorological Observatory, China Meteorological Administration (CMA), where it has been continuously running for over a year in a real-time forecasting environment. Throughout this period, it has consistently outperformed all other AI-based forecasting systems in use, delivering more accurate and reliable TC predictions that directly support emergency decision-making, public warnings, and maritime safety operations. This successful deployment represents a significant milestone in the operational integration of deep learning into high-stakes meteorological applications, demonstrating that AI models, when carefully designed with physical awareness and sub-grid-scale refinement, can meet the rigorous demands of real-world weather services.

The main contributions of this work are as follows:
\begin{enumerate}
    \item Our analysis reveals fundamental data-centric limitations that lead to the degraded performance of existing AI-based global weather models during intensity forecasting.
    \item We introduce BaguanCyclone, a framework designed to refine both track and intensity forecasting. It significantly outperforms existing AI-based models, \textbf{achieving a 16\% reduction in tracking error and a 34\% improvement in intensity precision}, respectively across all 6 regions in comparison with the vanilla model based on global weather foundation model.
    \item By resolving the non-linear feedbacks between internal TC processes and the environment, BaguanCyclone maintains high predictive skill across complex synoptic regimes, such as re-intensification, multi-TC, and meandering tracks, that often elude traditional numerical TC prediction.
    \item BaguanCyclone have been operationally deployed at the Zhejiang Meteorological Observatory, CMA during the 2025 WP typhoon season, achieving average track and intensity errors of 85.61 km and 6.20 m/s. During Typhoon Co-May in July, 2025, forecasts have supported the evacuation of 97,000 people from vulnerable coastal and low-lying regions in Zhejiang.
\end{enumerate}

\section{Related Work}
\subsection{Challenges in Typhoon Prediction}
Tropical cyclone is a rapidly rotating system characterized by a low-pressure center, closed low-level atmospheric circulation, strong winds, and a spiral arrangement of thunderstorms that generate torrential rain and squalls. While operational TC forecasting has seen steady progress in recent years, significant challenges remain. First, the prediction of rapid intensification (RI) event—defined as an increase in TC intensity of at least 30 knots (approximately 15.4 m/s) within a 24-hour period~\cite{kaplan2003large}—has proven especially difficult~\cite{kaplan2010revised, demaria2021operational, cangialosi2020recent}. This difficulty primarily stems from a limited understanding of the physical mechanisms underlying such extreme events~\cite{kaplan2003large}. Second, as a TC dissipates into a tropical depression (maximum sustained wind speed  < 17.1 m/s or 34 knots, according to diffirent standards~\cite{knapp2010international}), its structural integrity tends to become loose and highly asymmetric, leading to significantly inflated errors in both intensity estimation and center localization. Notably, the rare instances where a system undergoes re-intensification back into a stronger system—an "intensity rebound"—frequently result in major forecast discrepancies~\cite{multi-chen2021boundary, deng2025new}. Additionally, the complexity of multi-TCs environments poses a significant predictive hurdle; even in the absence of direct binary interactions (Fujiwhara effect~\cite{Fujiwhara}), the competitive depletion of ambient moisture and energy between coexisting systems can severely degrade model fidelity. Finally, the synergistic evolution between mid-latitude westly troughs and large-scale circulation patterns often induces erratic trajectories, further amplifying the prediction model inherent non-linearity and uncertainty~\cite{multi-nguyen2015simulation, multi-chen2021boundary, multi-alvey2022weak}. In summary, achieving the precise prediction of typhoon intensity evolution and movement trajectories remains a persistent and cutting-edge challenge in atmospheric science.

\subsection{Deep Learning for Tropical Cyclones}
Global weather foundation models (GFMs), such as FourCastNet~\cite{pathak2022fourcastnet}, Pangu-Weather~\cite{pangu_nature}, FuXi~\cite{Fuxi_nature}, GraphCast~\cite{graphcast}, and Baguan~\cite{niu2025utilizing}, have revolutionized TC track forecasting, with many now consistently surpassing the operational HRES benchmark. However, intensity prediction remains a persistent ``Achilles' heel'' for these systems. Recent benchmarking~\cite{huang2025benchmark, demaria2024evaluation} reveals that GFMs systematically underpredict peak wind speeds, often performing worse than rudimentary statistical baselines. This performance gap stems from the difficulty of capturing high-resolution structural nuances and physical consistency within global-scale architectures.
Concurrently, specialized deep learning models—such as ConvLSTM~\cite{shi2015convolutional}, VQLTI~\cite{wang2025vqlti}, and Deep-Hurricane-Tracker~\cite{kim2019deep}, have been developed to address TC structural evolution by incorporating physical constraints or multi-source data. Yet, a critical limitation persists: these specialized methods typically function as standalone systems, failing to leverage the robust, large-scale atmospheric representations inherent in GFMs.
To bridge this divide, We propose a Baguan-based refinement framework that synergizes global forecasting power with targeted optimizations for superior track and intensity fidelity.

\section{Methodology}
\begin{figure*}[t]
    \centering
    \includegraphics[width=0.8\textwidth]{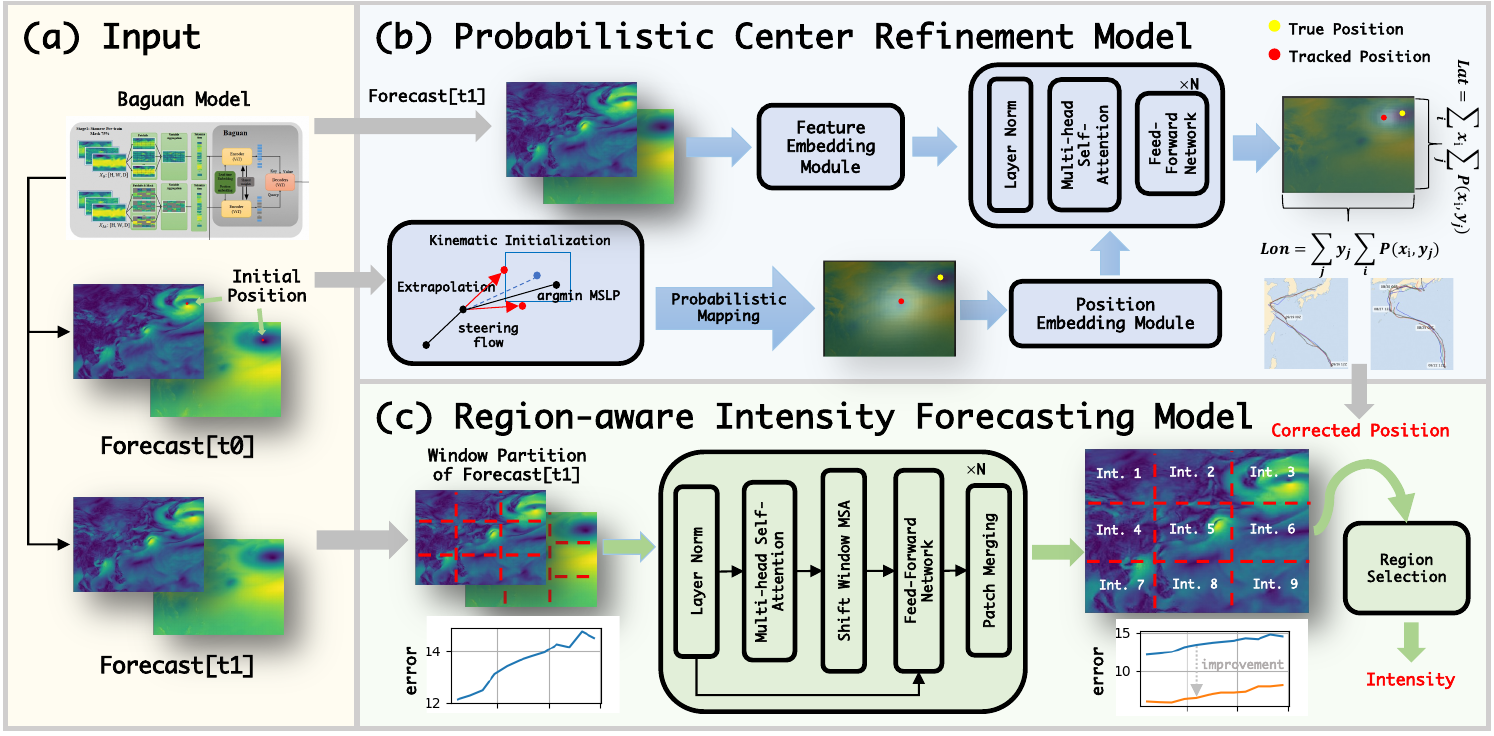}
    \caption{Overview of BaguanCyclone’s architecture. (a) The input consists of an initial position and 2 subsequent forecast frames. (b) The probabilistic center refinement model first employs a nonparametric algorithm to obtain an initial tracking distribution, which is then refined using a tracking correction model. (c) Region-aware intensity forecasting model leverages a Swin-transformer with patch merging to compute the per-region intensity.}
    \label{fig:arch}
    \vspace{-1em}
\end{figure*}

The architecture of BaguanCyclone, as shown in Fig.~\ref{fig:arch}, integrates capabilities for TC track prediction and intensity forecasting.
The \textbf{Probabilistic center refinement model} is designed to predict TC latitude and longitude coordinates with arbitrary precision by modeling location as a probability distribution map, using the initial position and a continuous sequence of atmospheric states of $0.25^\circ$ generated by an AI model as input.
The \textbf{region-aware intensity forecasting model} partitions the atmospheric states into multiple regions and then predicts the maximum wind speed for each region.
During inference, these two modules are coupled, with the predicted latitude and longitude localized within the intensity forecasting regions to retrieve the corresponding intensity information.

\subsection{Probabilistic Center Refinement Model}



Our methodology \textbf{transitions from a deterministic tracking paradigm to a probabilistic density learning framework}, effectively bypassing the spatial quantization (grid-locking) inherent in the $0.25^\circ$ reanalysis data. This process is performed in three stages: kinematic initialization, density mapping, and neural correction. By optimizing a softened spatial distribution via KL-divergence and decoding it through an expectation operator, the model achieves sub-grid localization. This allows for trajectory resolution with arbitrary precision, transcending the native resolution of the underlying atmospheric grid.

\subsubsection{Kinematic Initialization}
To establish a robust physical anchor, we first generate a deterministic trajectory prior by integrating a multi-step procedural tracker~\cite{Bodnar2024AuroraAF} with steering flow vectors from ECMWF~\cite{ECtracker}. An initial position estimate is obtained via the kinematic extrapolation of the historical trajectory, weighted by the local steering flow. This candidate position is then iteratively refined within a bounding box, which progressively narrows to converge upon the nearest local pressure minimum. While this provides a physically consistent estimate $(\hat{x}_c, \hat{y}_c)$, it remains constrained by the discrete grid resolution of the input fields.

\subsubsection{Probabilistic Density Mapping}
To represent spatial uncertainty and enable sub-grid refinement, the discrete coordinates are transformed into a continuous probability density field. Given a region of size $H \times W$, we apply a truncated Gaussian kernel to map the cyclone center into a softened spatial representation. The local weights $\tilde{g}_{ij}$ are computed based on the Euclidean distance $d_{ij}$ between the grid point $(i,j)$ and the center, subject to a fixed truncation radius $r$:
\begin{equation}
    \tilde{g}_{ij} = 
    \begin{cases}
        \mathbf{exp}(-\frac{d_{ij} ^ 2}{2\sigma^2}), & \text{if } d_{ij} \leq r, \\
        0, & \text{otherwise}.
    \end{cases}
\end{equation}
To ensure that the density field represents a valid probability mass, the weights are normalized:
\begin{equation}
    w_{ij} = \frac{\tilde{g}_{ij}}{\sum_{k}\sum_{l}\tilde{g}_{kl}}.
\end{equation}
This probabilistic encoding transforms a single point into a field of spatial influence, allowing the deep learning model to perceive the center's location relative to its surrounding atmospheric environment.



\subsubsection{Neural Correction and Reconstruction}
The tracking correction model ingests this probability field as an auxiliary spatial channel, concatenated with the multi-level atmospheric state features. This allows the model to learn the non-linear relationship between local meteorological dynamics and systematic tracking biases. The model outputs a predicted density field $\hat{W}$, which is normalized via a softmax operation to maintain the property $\sum_{ij}\hat{W}_{ij}=1$.
The optimization is guided by the KL divergence between the predicted distribution $\hat{W}$ and the ground-truth distribution $W_{gt}$:
\begin{equation}
    \mathcal{L}_{KL} = D_{KL}(W_{gt}||\hat{W}).
\end{equation}
Finally, to recover high-precision coordinates, we apply an expect\-ation-based reconstruction. The final latitude and longitude are calculated by computing the expected values along the spatial axes of the density field. By leveraging the continuous nature of the predicted probability mass, this mechanism produces coordinates that are no longer confined to the $0.25^\circ$ grid intervals, thereby achieving sub-grid localization and superior tracking fidelity.



\subsection{Why AI Models Failed in Intensity Forecasting: A Data-centric Perspective}

The majority of AI models in the literature report tracking accuracy only~\cite{pangu_nature, niu2025utilizing}, omitting an assessment of intensity forecasting performance, and there is growing evidence~\cite{huang2025benchmark} that such models exhibit significant limitations in predicting storm intensity.

\begin{table}[h]
\centering
\caption{Comparison of ERA5 ($0.25^\circ$) and EC Analysis ($0.1^\circ$) over the Western North Pacific (WP) and North Atlantic (NA). Results are presented only for 6-hour.}
\label{tab:tracking_gt}
\resizebox{0.95\columnwidth}{!}{
\begin{tabular}{c|cc|cc}
\toprule
& \multicolumn{2}{c|}{dist. (km)} & \multicolumn{2}{c}{wind (m/s)} \\ 
 & ERA5 ($0.25^\circ$) & EC Analysis ($0.1^\circ$) &  ERA5 ($0.25^\circ$) & EC Analysis ($0.1^\circ$) \\
\midrule
WP & 45.3 & 56.9 & 7.74 & 5.69 \\
NA & 32.7 & 48.1 & 11.53 & 8.87 \\
\bottomrule

\end{tabular}
}

\end{table}

Before introducing our intensity forecasting model, we first examine, from a data-centric perspective, the reasons underlying the poor performance of existing AI models in tropical cyclone track forecasting. We utilize the widely used ground truth data in the training of many AI models, the $0.25^\circ$ ERA5 reanalysis and the $0.1^\circ$ EC analysis. As shown in Tab.~\ref{tab:tracking_gt}, using the nonparametric tracking algorithm, we find that 
differences in spatial resolution lead to intensity forecasts that diverge by approximately 20\%, whereas their impact on track forecasts is considerably smaller.
Owing to its grid-averaged nature, ERA5 data is inherently limited in its ability to resolve extreme values—such as the peak intensity of tropical cyclones. As a result, AI models trained on ERA5 consistently underperform in intensity forecasting relative to NWP systems initialized with higher-resolution 0.1° analyses.

\subsection{Region-aware Intensity Forecasting Model}
Additionally, AI-based models are typically trained with MSE loss, which inherently biases predictions toward the conditional mean, further leading to suboptimal performance on extreme-value tasks.

\begin{table}[h]
\centering
\caption{Minimum inter-TC distance ($^\circ$) within each region from 2000 to 2023. A value of –1 indicates no twin TCs occurred.}
\label{tab:min_dis}

\resizebox{0.8\linewidth}{!}{
\begin{tabular}{c|cccccc}
\toprule
Basin & WP & EP & NA & NI & SI & SP \\
\midrule
Dist. & $3.37^\circ$ & $1.20^\circ$ & -1 & $16.37^\circ$ & $4.70^\circ$ & $5.92^\circ$ \\
\bottomrule

\end{tabular}
}

\end{table}

To this end, as shown in Fig.~\ref{fig:arch} (c), we propose the region-aware algorithm based on patch merging, a mechanism that naturally partitions the input into multiple regions. To better determine the region partitioning size, we conducted a statistical analysis of the minimum inter-TC distance across all regions, as summarized in Tab.~\ref{tab:min_dis}. Each region is therefore designed to have a size smaller than 1/3 of this minimum distance.
Specifically, the input size of $V \times H \times W$ is transformed into size of $\frac{H}{2^N \cdot p} \times \frac{W}{2^N \cdot p}$, where $p$ denotes the patch size used in the patch embedding layer, $N$ is the number of patch merging blocks and the first dimension.

\subsection{Joint Deployment of Tropical Cyclone Forecasting Models at Inference}

In the inference phase, our framework employs a dual-stream architecture that jointly deploys the probabilistic center refinement model alongside the region-aware intensity forecasting model. Upon receiving the input atmospheric states, the system processes the data through both channels concurrently: the center refinement module generates the predicted coordinates of the cyclone’s eye, while the intensity forecasting module produces a comprehensive map of predicted maximum intensities across various spatial partitions. To synthesize these outputs, the predicted center coordinates are utilized as a spatial query to index into the partitioned intensity map. By identifying the specific region that encompasses the refined center, the corresponding intensity value is extracted and designated as the final, authoritative intensity prediction for the cyclone.

\section{Experiments}

\begin{figure*}[h]
    \centering
    \includegraphics[width=0.85\textwidth]{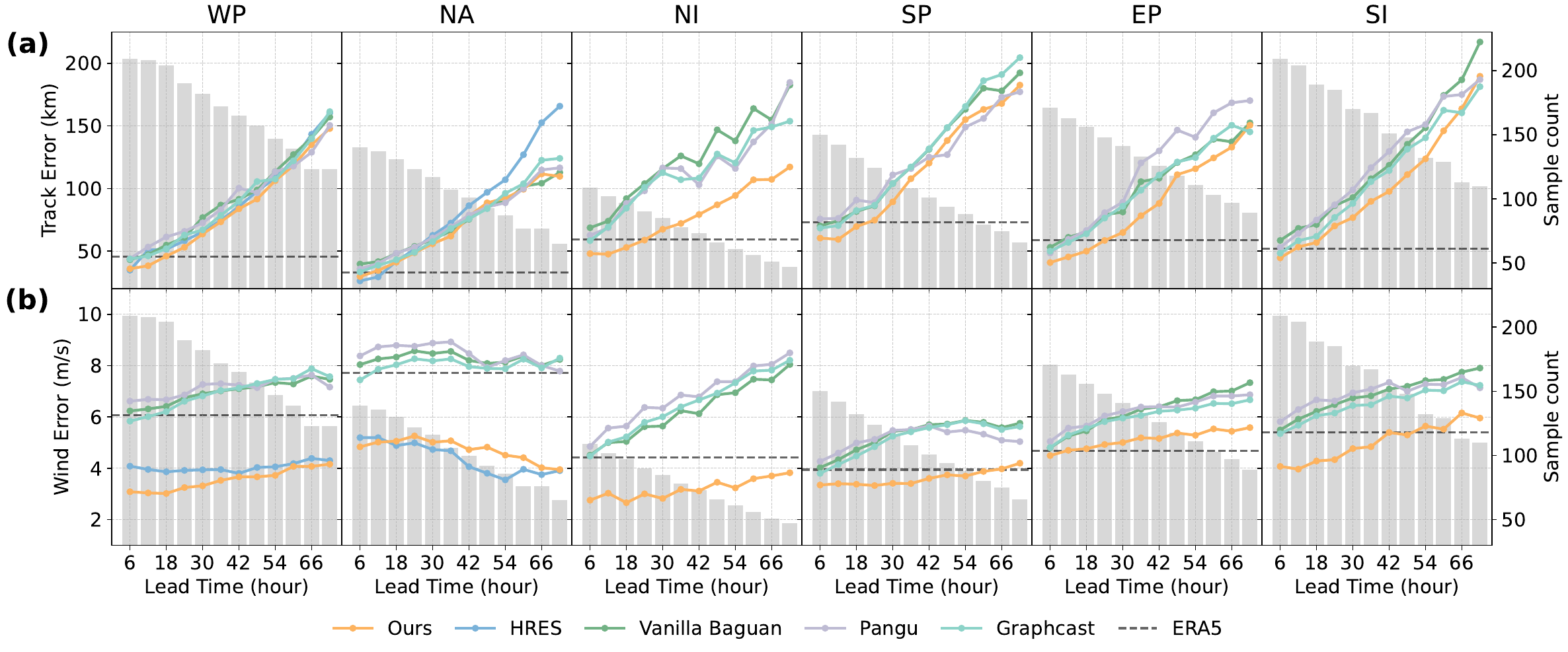}
    \Description{A three-panel figure...}
    \caption{Bias in track and intensity predictions. Solid lines correspond to forecast models:BaguanCyclone (ours),  HRES (ECMWF), Vanilla Baguan, Pangu-Weather, and Graphcast. As ERA5 (the horizontal dashed lines) represents reanalysis data rather than predictive forecasts, its 6-hour intrinsic error is depicted as a constant baseline. The gray histograms indicate the sample count of the validation set.}
    \label{fig:maintable}
\end{figure*}

\subsection{Settings}

We utilize IBTrACS~\cite{knapp2010international} as the ground-truth best-track repository. Training features are derived from ERA5 reanalysis, while Baguan forecasts~\cite{niu2025utilizing} are employed during evaluation to simulate operational conditions. Input features comprise atmospheric variables (Z, T, U, V, W, Q) at three pressure levels (500, 700, 850 hPa), supplemented by surface-level parameters including 2m temperature/dewpoint, 10m wind components, mean sea level (MSL), surface pressure, and total column water/vapor. Due to limited 2024 samples in the NI and SP basins, the training and evaluation periods are set to 2000–2022 and 2023–2024, respectively. For all other regions, the model is trained on 2000–2023 and evaluated on 2024.
The tracking performance is measured by the great-circle distance (km) between forecast and IBTrACS coordinates using the Haversine formula. Intensity is evaluated via the mean absolute error (m/s) of the maximum wind speed. Detailed descriptions of the datasets and evaluation metrics are provided in App.~\ref{app:setting}.

\subsection{Main Results}
\subsubsection{Baselines} 

We employ the most recent AI-based model, Ba\-guan, to generate atmospheric states for evaluation. To enable a meaningful and fair comparison, we select both NWP and AI-based methods. For the NWP method, we select the leading ECMWF \textbf{HRES}. In terms of AI-based methods, we select those methods based on global weather foundation models due to their outstanding performance, including \textbf{Vanilla Baguan} (without the refinement module), \textbf{Pangu-Weather} and \textbf{GraphCast}. We evaluate performance across six major global tropical cyclone basins (WP, EP, NA, NI, SI and SP). Furthermore, to contextualize the performance of BaguanCyclone relative to reanalysis quality, we also visualize the intrinsic error of the ERA5 dataset at a 6-hour lead time. The SA basin was not included in this study due to the scarcity of historical samples required for robust model training. 

Since TC forecasts within a 3-day lead time are of greater practical significance, we focus our comparison on this period. 
Furthermore, the HRES only provides forecasts when the initial intensity exceeds a specific threshold~\cite{richardson2012new}, leading to insufficiant data samples in certain regions. Thus, for the WP and NA regions, the evaluation is based on data samples where HRES forecasts are successfully matched with IBTrACS observations. 
In other regions, the comparison is limited to AI models and excludes HRES, as these models provide comprehensive coverage for all cases documented in the IBTrACS.

\subsubsection{Tracking}
As illustrated in Fig.~\ref{fig:maintable} (a), BaguanCyclone delivers superior tracking performance across all six regions. Notably, BaguanCyclone consistently surpasses the HRES benchmark across the majority of forecast lead times. This achievement is particularly significant as it maintains a competitive edge despite the inherent constraints of a relatively coarse $0.25^\circ$ input resolution, underscoring the model's exceptional architectural efficiency. By integrating our proposed methodology, BaguanCyclone yields an average improvement of 16\% over all AI-based baselines. Specifically, within the highly active WP basin, the model demonstrates significant gains over Vanilla Baguan, achieving a 16.14\% error reduction in 24-hour short-range forecasts and an 11.3\% improvement at the 72-hour long-range horizon. Performance gains are even more prominent in the NI basin, where BaguanCyclone achieves a substantial reduction in tracking error, exceeding 30\%, relative to AI-based baselines.

\subsubsection{Intensity Prediction} 
As illustrated in Fig.~\ref{fig:maintable} (b), conventional AI-based baselines (including Vanilla Baguan) exhibit a persistent deficiency in intensity forecasting. This limitation, well-documented in existing literature~\cite{huang2025benchmark, zhong2024fuxi}, is further exacerbated by the inherent intensity biases within the ERA5 reanalysis dataset, as visualized in the Fig.~\ref{fig:maintable} (b). While such data-quality bottlenecks typically constrain the performance of data-driven frameworks, BaguanCyclone overcomes these limitations, delivering substantial gains that surpass the HRES benchmark across most lead times. This achievement effectively bridges the gap between reanalysis-trained research models and stringent operational standards. Notably, in the WP basin, our model consistently outperforms HRES across all lead times by an average margin of 12.23\%. 
Compared with AI-based baselines, BaguanCyclone demonstrates a substantial performance leap, yielding a 34\% average reduction in forecast error. Notably, this improvement reaches a remarkable 50\% in the NI basin, demonstrating BaguanCyclone’s profound utility for real-world meteorological applications.

 
\subsection{Generalization Capabilities across Various Tropical Cyclone Varieties}
\label{sec:tc-varieties}

Tropical cyclones (TCs) exhibit diverse track and intensity evolutions controlled by different mechanisms, including steering flow,
land--sea energy-budget transitions, and nonlinear interactions between TC and background fields. Rather than treating these as isolated ``special cases'', we use
representative events to test the following science-oriented hypothesis:

\smallskip
\noindent\textbf{Hypothesis (AI4Science).}
\emph{BaguanCyclone generalizes across TC varieties by correcting two key failure modes of coarse-resolution AI forecasts:
(i) discretization-induced center errors and (ii) core-intensity underestimation, thereby remaining consistent with the governing drivers
(steering flow, surface energy supply, and synoptic competition) even when the dominant regime shifts.}
\smallskip

Across the cases below, BaguanCyclone's gains are not only numerical; they reflect mechanism-aware robustness under three recurring drivers:
(1) steering-flow guidance, (2) land--sea regime transitions, and (3) nonlinear competition under weak or interacting systems.

\subsubsection{Steering-flow guidance under sweeping-arc tracks (Beryl)}
\label{subsec:beryl}
\begin{figure}[h]
    \centering
    \includegraphics[width=1\linewidth]{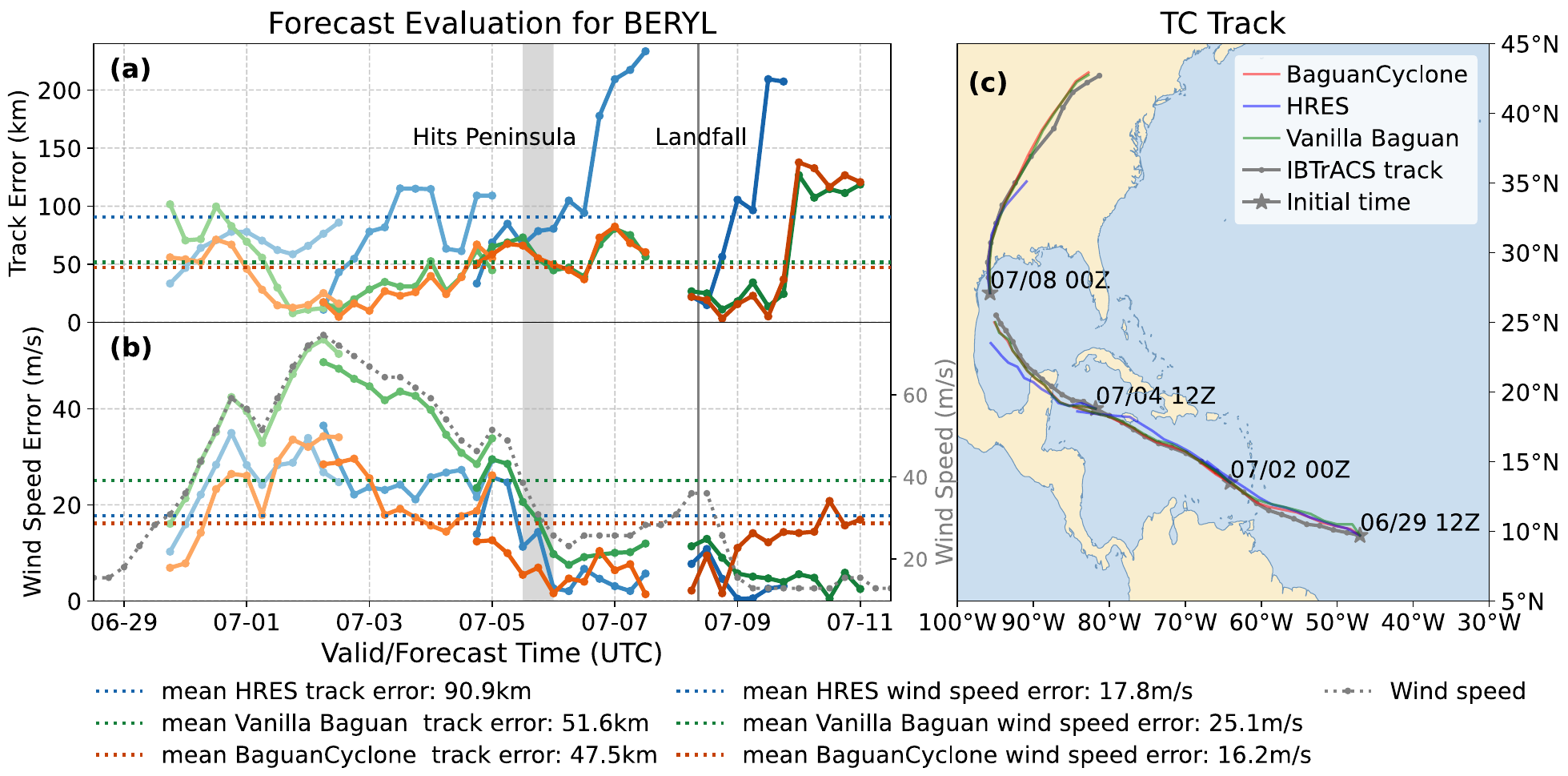}
    \Description{A three-panel figure...}
    \caption{Forecast evaluation for tropical cyclone Beryl: (a) track error, (b) max wind error, and (c) observation vs. prediction. Black line and gray shading indicate landfall and re-emergence; dashed line denotes mean error.
    }
    \label{fig:beryl}
    \vspace{-1em}
\end{figure}

Many TCs follow sweeping-arc tracks (including recurvature), primarily controlled by large-scale steering flow from subtropical highs and
midlatitude troughs. In this regime, small center errors on coarse grids can amplify rapidly, especially near turning points.
We examine TC Beryl (Fig.~\ref{fig:beryl}), rapid intensification (RI) event, weakening, land interaction, and post-landfall evolution. Incorporating physical guidance from meteorological fields into the learning of nonlinear rules, BaguanCyclone provides more accurate track and intensity forecasts over most of the 72-hour window initialized at 18:00~UTC on 29~June, indicating improved robustness for intensification-related evolution.

Beryl's track was strongly influenced by the Azores (Bermuda) High \cite{Hordon2005}. BaguanCyclone captures this control by maintaining
multi-level dynamical consistency across 500/700/850~hPa and surface winds, stabilizing the inferred steering signal through recurvature and landfall.
It also responds to land transition via storm-relative humidity and low-level wind gradients consistent with increased surface friction and associated
kinetic energy dissipation.

\subsubsection{Regime shifts in intensity: landfall weakening and re-intensification (Pulasan)}
\label{subsec:pulasan}
\begin{figure}[h]
    \centering
    \includegraphics[width=1\linewidth]{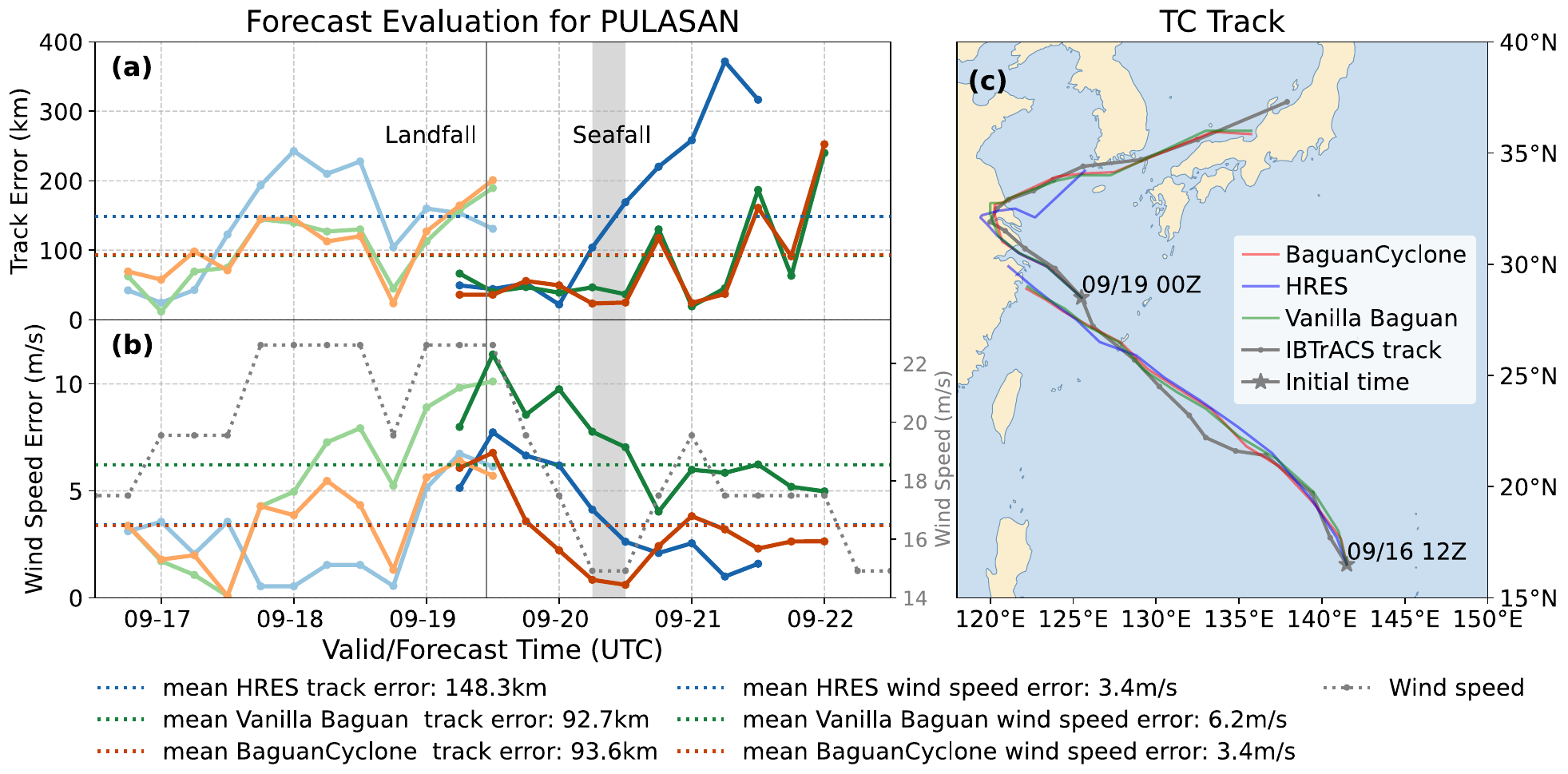}
    \Description{A three-panel figure...}
    \caption{Forecast evaluation for tropical cyclone Pulasan, following the same format as Fig.~\ref{fig:beryl}.}
    \label{fig:pulasan}
\end{figure}

Intensity forecasting is particularly difficult for AI models trained on coarse reanalysis due to smoothed core extremes at 0.25$^\circ$ and
regression-to-mean effects. We evaluate TC Pulasan (Fig.~\ref{fig:pulasan}), which weakened after landfall and re-intensified after returning to the ocean.

Landfall largely cuts off oceanic heat/moisture fluxes and increases surface friction, often yielding a negative energy budget and rapid decay.
Pulasan, however, remained over land briefly and retained structural coherence, enabling re-intensification. BaguanCyclone anticipates the rebound
over the Yellow Sea under supportive thermodynamic conditions (warm underlying and abundant moisture), suggesting improved generalization across land--sea
transitions by anchoring intensity estimation to the storm-centered local environment rather than coarse-grid smoothing.

\subsubsection{Nonlinear interactions: multiple cyclones and rapid steering reversals (Trami and Kong-rey)}
\label{subsec:twins}
\begin{figure}[h]
    \centering
    \includegraphics[width=0.9\linewidth]{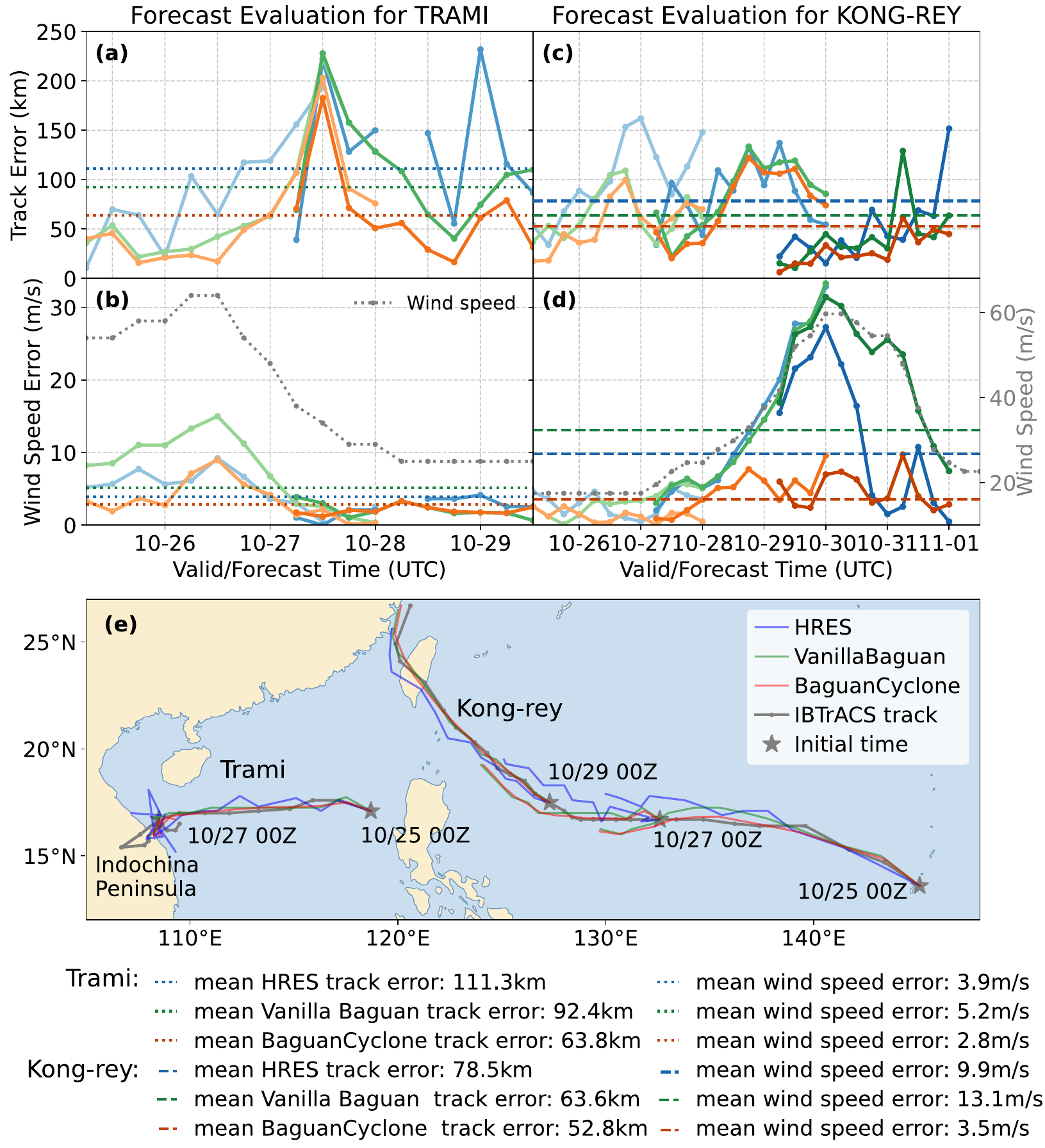}
    \Description{A three-panel figure...}
    \caption{Forecast evaluation for tropical cyclone Trami and Kong-rey, following the same format as Fig.~\ref{fig:beryl}.}
    \label{fig:twins}
\end{figure}

Multiple or closely spaced cyclones require maintaining cyclone identity while disentangling storm-relative dynamics under shared environmental flow.
We analyze the concurrent Trami and Kong-rey event (Fig.~\ref{fig:twins}). BaguanCyclone decouples their steering environments from multi-level fields,
yielding a more accurate northwestward track for Kong-rey under the subtropical high. It also better captures peak intensity over the open ocean,
with lower errors near maximum winds.

This case further includes Trami's rare $180^\circ$ post-landfall reversal over the Indochina Peninsula. Such U-turns are typically tied to rapid
adjustments of the midlatitude trough and subtropical high that flip the steering flow. BaguanCyclone reproduces the reversal, indicating sensitivity
to fast synoptic transitions rather than regression-to-mean behavior.

\subsubsection{Weak-steering environments: meandering and stalling trajectories (Shanshan)}
\label{subsec:shanshan}
\begin{figure}[h]
    \centering
    \includegraphics[width=1\linewidth]{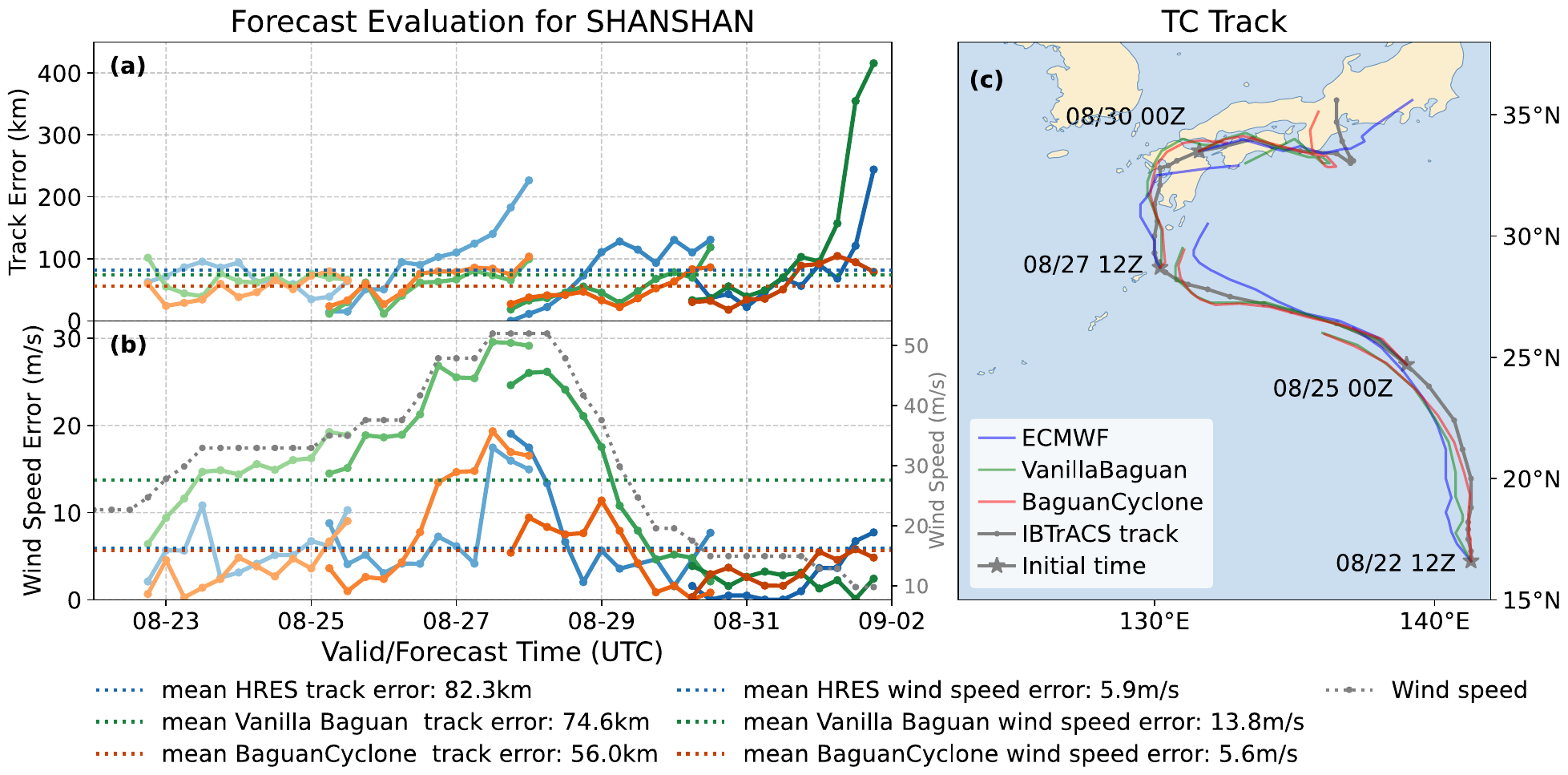}
    \Description{A three-panel figure...}
    \caption{Forecast evaluation for tropical cyclone Shanshan, following the same format as Fig.~\ref{fig:beryl}.}
    \label{fig:shanshan}
\end{figure}

When steering flow is weak, TC motion becomes highly nonlinear and sensitive to subtle circulation gradients, often leading to meandering and stalling tracks.
We evaluate TC Shanshan (Fig.~\ref{fig:shanshan}), which exhibited weak-steering meandering and later abrupt deflections.

BaguanCyclone captures the slow, erratic motion by responding to circulation gradients that reflect the competition between the subtropical high and
midlatitude westerly troughs. After landfall, it remains robust during dissipation, reproducing the abrupt, nearly $90^\circ$ deflection in the terminal stage,
consistent with interactions between the decaying vortex and an approaching midlatitude trough.

\paragraph{Summary (AI4Science).}
Across sweeping arcs, re-intensification, multiple cyclones, and meandering events, BaguanCyclone remains skillful as the dominant driver shifts:
(i) steering flow control, (ii) surface energy-budget transitions, and (iii) nonlinear competition under weak or interacting systems.
These cases support systematic bias correction as a mechanism-aware refinement layer that yields more physically coherent TC track and intensity evolutions
from coarse-resolution AI forecasts.

\subsection{Ablations}


\subsubsection{Probabilistic Forecasting vs. Residual Forecasting}

To validate the design choice of the probabilistic model for TC center locations, we compare our probabilistic formulation against a deterministic variant that directly forecasts the residual of latitude and longitude coordinates using the same backbone architecture. Both models are trained under identical conditions and evaluated on the same test set.

As shown in Tab.~\ref{tab:ablation_prob}, the probabilistic approach mostly outperforms the direct regression baseline in all evaluation metrics, particularly in terms of localization accuracy and robustness to trajectory outliers. More importantly, the probabilistic formulation provides calibrated uncertainty estimates, enabling risk-aware decision-making. For example, high predictive entropy correlates strongly with challenging forecasting scenarios such as rapid direction shifts or landfall events. In contrast, the deterministic model offers no measure of confidence, and tends to produce overconfident yet inaccurate predictions under distributional shift. These results underscore the value of explicit uncertainty modeling in geospatial forecasting tasks where prediction reliability is as critical as point accuracy.

\begin{table}[h]
\centering
\caption{A comparative analysis of probabilistic and residual forecasting in the tracking refinement module in WP and NA. The error is measured in km. Lower value is better.}
\label{tab:ablation_prob}

\resizebox{0.8\linewidth}{!}{
\begin{tabular}{c|ccc|ccc}
\toprule
Lead & \multicolumn{3}{c|}{WP} & \multicolumn{3}{c}{NA} \\
Time & Vanilla & Prob. & Res. & Vanilla & Prob. & Res. \\
\midrule
24h & 60.9 & \textbf{53.0} & 60.7& 53.8 & \textbf{48.6} & 49.8 \\
48h & 98.8& \textbf{91.5} & 101.1& 92.8& 88.5 & \textbf{87.6} \\
72h & 157.1 & \textbf{147.9} & 158.4& 112.8 & \textbf{109.6} & 111.4 \\
\bottomrule

\end{tabular}
}
\end{table}

\subsubsection{Coupling Tracking and Intensity Module}

To evaluate the effectiveness of the coupled architecture, we conducted a comparative analysis by decoupling the track and intensity modules. Specifically, we used both the vanilla Baguan predicted tracks and the corrected tracks as inputs to the intensity module for post-hoc intensity estimation. These results serve as baselines for quantifying the performance gains achieved by our coupled modeling approach. As shown in Tab.~\ref{tab:ablation_coupling}, our experiments in the  SI region demonstrate that the coupled architecture yields significant improvements in the vast majority of cases.
This improvement can be attributed to the fact that the refined tracks provide more precise locations of the TCs within the spatial window. Consequently, this reduces intensity forecast biases that typically stem from track displacement errors, ensuring the intensity module extracts features from the correct atmospheric environment

\begin{table}[h]
\centering
\caption{Performance comparison of coupled vs. separate SI models. Values indicate \% improvement; positive values favor the coupled model.}
\label{tab:ablation_coupling}

\resizebox{0.85\linewidth}{!}{
\begin{tabular}{c|cccccc}
\toprule
Lead & 12h & 24h & 36h & 48h & 60h & 72h \\
\midrule
wind (m/s) & +0.89 & +1.50 & +2.40 & +2.88 & +3.74 & +0.15 \\
pres (hpa) & +1.96 & +0.82 & +1.87 & +0.81 & +1.81 & -0.55 \\
\bottomrule

\end{tabular}
}

\end{table}

\subsection{Real-world Deployment}

In July 2025, BaguanCyclone is officially operationalized at the Zhejiang Meteorological Observatory, CMA. By delivering high-precision trajectory predictions for all WP typhoons, the system provided critical decision support for regional disaster mitigation. In July 2025, the operational value of BaguanCyclone is demonstrated through its accurate predictions of Typhoon Co-May. \textbf{By providing reliable landfall and intensity data, the model supported the organized evacuation of 97,000 people from vulnerable coastal and low-lying regions in Zhejiang}.

\begin{figure}[h]
    \centering
    \includegraphics[width=0.9\linewidth]{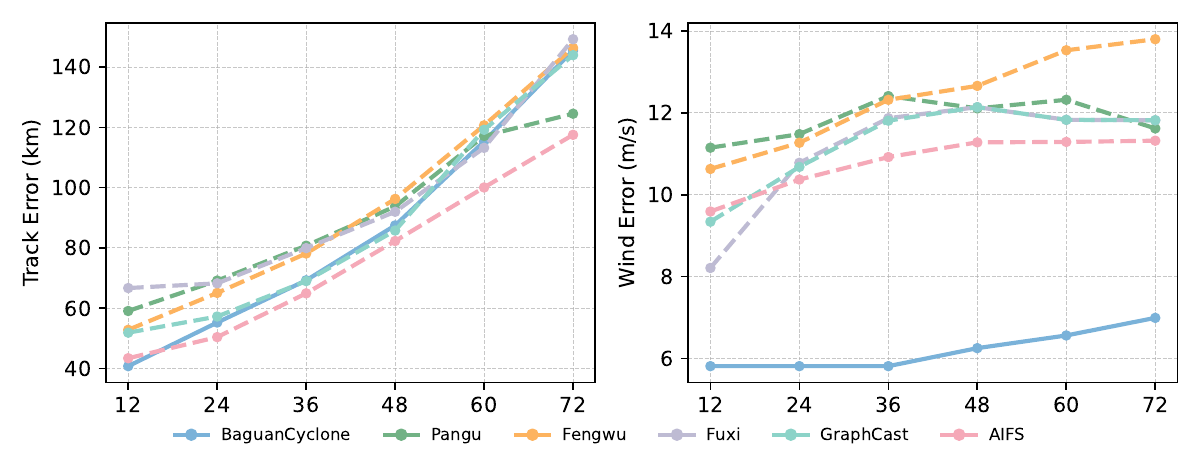}
    \caption{Average real-time forecasting performance during the 2025 WP typhoon season. Note that most baseline models (e.g., FuXi and Pangu-Weather) are not the latest versions.}
    \label{fig:deploy}
\end{figure}

Fig. \ref{fig:deploy} presents the operational performance of BaguanCyclone alongside several AI-based models, including FuXi~\cite{Fuxi_nature}, Pangu-Wea\-ther~\cite{pangu_nature}, GraphCast~\cite{graphcast}, FengWu~\cite{Chen2023FengWuPT}, and AIFS~\cite{lang2024aifsecmwfsdatadriven}. While these models are operational at the CMA, it should be noted that their performance may reflect specific deployed versions rather than the latest iterations. Among these, BaguanCyclone demonstrates clear superiority across both key metrics. It leads the majority of models in track forecasting with a mean error of 85.61 km. Most notably, its performance in intensity prediction represents a landmark achievement—its average error of 6.20 m/s constitutes an unprecedented 50\% reduction compared to the second-best performer. This paradigm-shifting precision effectively overcomes long-standing bottlenecks in intensity forecasting, offering a level of reliability previously considered unattainable in operational settings.

\section{Conclusion}
In this study, we proposed a novel AI-based modeling framework specifically designed for TC's track and intensity forecasting. Our models have achieved substantial performance gains over existing AI-based weather forecasting models. Beyond theoretical validation, the proposed system has been successfully deployed at the Zhejiang Meteorological Observatory, CMA, for operational use. Over a continuous one-year evaluation period, the system demonstrated exceptional stability and predictive accuracy for typhoon events in the WP basin, significantly outperforming baseline AI models in real-world scenarios. This work bridges the gap between experimental AI research and practical meteorological operations.

\section{Limitations and Ethical Considerations}
This study uses only gridded meteorological/geophysical datasets, with no personal data or human subjects; all data licenses are followed. 
There are two limitations: the need for higher spatial resolution ($0.1^\circ$) to resolve finer atmospheric features, and the inherent `black-box' nature of the AI architecture. Future efforts will focus on bridging these gaps by integrating higher-granularity data and exploring the physical drivers behind the model’s decisions to improve both precision and transparency.

\clearpage

\section*{GenAI Disclosure}
During the preparation of this work the authors used GenAI in order to improve language. After using this tool, the authors re-viewed and edited the content as needed and take full responsibility for the content.


\bibliographystyle{ACM-Reference-Format}
\bibliography{mybib}

\appendix
\section{Appendix}

\subsection{Settings}
\label{app:setting}

\subsubsection{Datasets}
IBTrACS~\cite{knapp2010international} serves as the ground truth, being the most comprehensive global repository of TC best-track data. It integrates recent and historical TC observations from multiple meteorological agencies into a unified, publicly available dataset.
The atmospheric state used during training is derived from ERA5 reanalysis data. For evaluation, to better reflect real-world operational conditions, we instead employ atmospheric forecasts from Baguan~\cite{niu2025utilizing}.

Our input features include 5 standard vertical pressure levels (500, 700, and 850 hPa), with key atmospheric variables at each level: geopotential, temperature, u-component of wind, v-component of wind, w-component of wind, and specific humidity. Additionally, we incorporate the following surface-level variables: 2-meter temperature (T2M), 2-meter dewpoint temperature (D2M), 10-meter wind components (U10, V10, W10), mean sea-level pressure (MSL), surface pressure (SP), total column water (TCW), and total column water vapor (TCVW).
For NI and SP, the model is trained on data from 2000–2022 and evaluated on 2023–2024. 
For other regions, the training period spans 2000–2023, with evaluation conducted on 2024 data.

\subsubsection{Metrics}
For track evaluation, we compute the great-circle distance (in kilometers) between forecasts and observations using the Haversine formula~\cite{1984S&T68R.158S}.
Given two geographic coordinates-$(\phi_1, \lambda_1)$ for the forecast location and $(\phi_2, \lambda_2)$ for the best-track observation, where $\phi$ denotes latitude and $\lambda$ denotes longitude (both expressed in radians)—the central angle $c$ between the points is calculated as:
\begin{equation}
    a = \mathrm{sin}^2(\frac{\Delta \phi}{2}) + \mathrm{cos}\phi_1\mathrm{cos}\phi_2\mathrm{sin}^2(\frac{\Delta \lambda}{2})
\end{equation}
\begin{equation}
    c = 2\mathrm{arctan2}(\sqrt{a}, \sqrt{1 - a}).
\end{equation}
where $\Delta \phi = \phi_2 - \phi_1$ and $\Delta \lambda = \lambda_2 - \lambda_1$. The corresponding surface distance $d$ is then obtained as:
\begin{equation}
    d = R \cdot c,
\end{equation}
with $R=6371\mathrm{km}$ representing the mean radius of the Earth.
All input coordinates are first converted from degrees to radians prior to computation.
For intensity evaluation, we use absolute error directly in m/s for maximum wind speed.

\end{document}